\documentclass{article} 
\usepackage{iclr2027_conference,times}
\usepackage{booktabs}
\usepackage{graphicx}
\usepackage{float}
\usepackage{wrapfig}
\usepackage{multicol}
\usepackage{algorithm}
\usepackage[noend]{algpseudocode}
\usepackage{multirow}
\usepackage{colortbl}
\usepackage{booktabs}
\usepackage{capt-of}
\usepackage{xcolor}
\usepackage[most]{tcolorbox}
\usepackage{pifont}

\definecolor{checkgreen}{HTML}{218739}
\definecolor{crossred}{HTML}{C63838}

\newcommand{\cmark}{\textcolor{checkgreen}{\ding{51}}}
\newcommand{\xmark}{\textcolor{crossred}{\ding{55}}}
\usepackage{amsmath,amssymb,booktabs,tabularx,pifont}

\usepackage[table]{xcolor}

\usepackage{amsmath,amsfonts,bm}

\def\eqref#1{equation~\ref{#1}}

\def\1{\bm{1}}

\DeclareMathAlphabet{\mathsfit}{\encodingdefault}{\sfdefault}{m}{sl}
\SetMathAlphabet{\mathsfit}{bold}{\encodingdefault}{\sfdefault}{bx}{n}

\usepackage{hyperref}
\usepackage{url}

\title{Order-Only On-Policy Distillation}

\author{
  Guangzeng Han \\
  University of Memphis \\
  \texttt{ghan@memphis.edu}
  \And
  Yu Wu \\
  University of Missouri \\
  \texttt{yw8g4@umsystem.edu}
  \And
  Xiaolei Huang \\
  University of Missouri \\
  \texttt{xiaolei.huang@missouri.edu}
}

\iclrfinalcopy 
\begin{document}

\maketitle
\fancyhead{}
\lhead{Preprint}
\renewcommand{\headrulewidth}{0.4pt}

\begin{abstract}
On-policy distillation (OPD) trains a student on its own generations
by matching the teacher's token probabilities. These probabilities
entangle two kinds of information: which tokens the teacher prefers,
and how confident it is. The latter reflects the teacher's own
capacity and calibration, which a smaller student may be unable to
reproduce. Because the teacher temperature changes these values
without changing the ordering, OPD is also sensitive to the
temperature choice. We ask whether the teacher's ordering of
candidate tokens is enough, and propose Order-Only On-Policy
Distillation (O$^{3}$PD). At each student-generated prefix, O$^{3}$PD
trains the student to reproduce the order of the teacher's top-$K$
tokens, without fitting their probability values. The objective is
therefore invariant to the teacher temperature and needs no
temperature tuning. Across six teacher--student pairs, O$^{3}$PD
outperforms OPD on the average over HMMT25, AIME24, and AIME25 in
every pair, by up to 2.9 points, with gains of up to 5.6 points on
AIME25. Other ranking losses yield similar gains, so the benefit
comes from ordinal supervision itself. Analysis shows that O$^{3}$PD departs further from the teacher's
probabilities than OPD, yet reproduces its ranking more faithfully
and yields a stronger student. The ranking, not the probability
values, thus carries the knowledge that on-policy distillation needs
to transfer.
\end{abstract}

\section{Introduction}
\label{sec:introduction}

Knowledge distillation (KD) transfers the capabilities of a strong
teacher model to a weaker student.
\citet{hinton2015KD} showed that the teacher's full output distribution,
rather than a single hard label, is what makes this transfer
effective: the probabilities assigned to competing outputs encode
relationships that a hard label cannot express. This insight made
distribution matching the default recipe for distillation. The
student is trained to reproduce not only which output the teacher
prefers, but also the exact probability that the teacher assigns to
every alternative.

\begin{wrapfigure}{r}{0.6\linewidth}
    \centering
    \includegraphics[width=\linewidth]{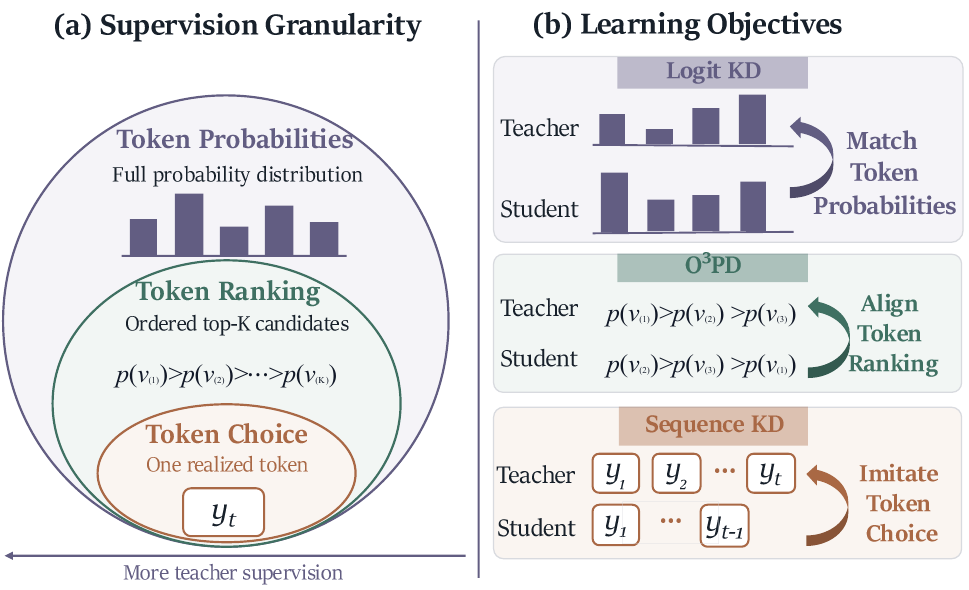}
    \caption{(a) Three nested levels of teacher supervision. (b) The corresponding objectives.}
    \label{fig:1}
\end{wrapfigure}

For autoregressive language models, the contexts in which
supervision is given matter as much as the supervision itself.
Sequence-level distillation trains the student with supervised
fine-tuning (SFT) on teacher-generated responses~\citep{SeqKD}. This
recipe has two weaknesses. Each position supplies a single target
token, so the teacher's preferences over alternatives are lost. In
addition, the student is trained on teacher-generated prefixes but
generates from its own prefixes at inference time, so its errors lead
it into contexts it has never been trained on. On-policy distillation
(OPD) removes this mismatch by querying the teacher on
student-generated trajectories and training the student to match the
teacher's token distribution at the states it actually
visits~\citep{MiniLLM,gkd}. OPD has since become a standard approach
for distilling reasoning models~\citep{yang2025qwen3technicalreport,OPSD}.

Matching probabilities, however, asks the student for more than it
needs. The teacher's probability values mix two kinds of
information: which tokens it prefers, and how confident it is in
that preference. The first is what the student must acquire to
reason like the teacher. The second reflects the teacher's own
capacity and calibration, which a much smaller student may be unable
to reproduce~\citep{EfficacyKD,mirzadeh2020improveddistillationVIATA,CalibrationKD}.
A distribution-matching objective penalizes every
deviation from these values, so the student spends much of its
capacity fitting a confidence profile that it cannot reach. The
problem shows up clearly in practice. OPD is highly sensitive to the
teacher's logit temperature, and the best temperature differs from
one teacher--student pair to another (Figure~\ref{fig:heat}).
Moreover, OPD can degrade the student as training continues, falling
below the base model it started from (Figure~\ref{fig:curves}). These
observations raise a basic question: \emph{does on-policy
distillation need the teacher's probabilities at all, or is the
teacher's ordering of candidate tokens enough?}

We answer this question by separating three nested levels of teacher
supervision: token choice, token ranking, and token probabilities
(Figure~\ref{fig:1}a). A sampled target reveals a single choice. A
ranked candidate list reveals which tokens the teacher considers
plausible and in what order it prefers them. A probability
distribution additionally fixes a numerical confidence for each
token. Existing objectives occupy the two ends of this hierarchy
(Figure~\ref{fig:1}b). Sequence KD imitates only the realized token,
while logit KD, including distribution-matching OPD, matches the full
probability distribution. Soft-target distillation shows that information about
alternatives matters, but it does not show that their exact
probabilities must be transferred. We argue that the middle level,
namely the teacher's ranking, carries the knowledge that matters for
distillation, and that the remaining probability values act as a
constraint rather than a signal.

Building on this view, we propose Order-Only On-Policy Distillation
(O$^{3}$PD). At each student-generated prefix, the teacher selects
its top-$K$ tokens and orders them by probability. The student is
then trained with an adjacent logistic ranking loss, which requires
each candidate to score higher than the next one in the teacher's
order. The teacher's probabilities decide which candidates are
selected and how they are ordered, but their values never enter the
loss. The $K-1$ adjacent constraints determine the complete ranking
through transitivity, so O$^{3}$PD transfers the full teacher order
with a minimal set of constraints while leaving the student free to
set its own probability values. Because dividing logits by a positive
temperature does not change their order, O$^{3}$PD is invariant to
the teacher temperature by construction and needs no temperature
tuning. Ordinal and relational supervision have been studied in
classification distillation and in learning to
rank~\citep{DIST,RankDistil,RKKD,OPKD,Plackett-Luce-Distillation},
but typically alongside KL terms or confidence weights, and over
fixed label sets with ground-truth labels. O$^{3}$PD uses the
teacher's order as its only signal and applies it to on-policy
distillation of autoregressive models, where the candidate list is
rebuilt by the teacher at every student-generated prefix.

We evaluate O$^{3}$PD on six Qwen3 teacher--student configurations
and three competition-level mathematics benchmarks: HMMT25, AIME24,
and AIME25. Against an OPD baseline, O$^{3}$PD improves the three-benchmark
average in all six configurations, by up to 2.9 points. It improves
AIME25 in every configuration, by as much as 5.6 points, and it
outperforms SFT and GRPO at every student scale. O$^{3}$PD also
trains stably: while OPD peaks early and then declines, O$^{3}$PD
keeps improving until the end of training. Our analysis explains
where these gains come from. O$^{3}$PD agrees less with the teacher's
probability values than OPD does, yet it reproduces the teacher's
token ordering and candidate set more faithfully. It also moves the
student much further from its initialization, most strongly for the
smallest student, which is also the student that gains the most.
Probability matching thus holds the student close to its starting
point, whereas ordinal supervision lets the student reshape its
distribution in the direction the teacher's preferences indicate.

Our contributions are as follows:
\begin{itemize}
    \item We identify probability matching as an unnecessary
    constraint in on-policy distillation. We separate the teacher's
    token ordering from its probability values and show that the
    ordering alone provides a stronger training signal.

    \item We introduce O$^{3}$PD, which trains the student  to reproduce the teacher's top-$K$
    token order, by default through adjacent pairwise constraints.
    O$^{3}$PD is invariant to the teacher temperature, requires no
    temperature tuning, and remains effective across different
    ranking objectives.

    \item We show that O$^{3}$PD outperforms OPD on the
    three-benchmark average in all six teacher--student
    configurations and exceeds SFT and GRPO in every configuration.
    Through analysis, we show that O$^{3}$PD learns the teacher's
    ranking rather than its probabilities, and that it allows the
    smallest student to move substantially further from its
    initialization.
\end{itemize}

\section{Related Work}
\label{sec:related}

Knowledge distillation (KD) can be characterized along two largely orthogonal axes: \emph{where} supervision is applied (the prefix states on which the student is trained) and \emph{what} teacher information is transferred at each state.

\subsection{LLM Distillation: From Teacher States to Student States}
\label{sec:related_llm_kd}

Early LLM distillation trains the student on fixed sequences. Sequence-level distillation uses teacher-generated responses as targets~\citep{SeqKD}, optionally enriched with rationales~\citep{DistillingStepByStep} or organized as preferences~\citep{dadpo}, while white-box variants match the teacher's token distributions along such sequences~\citep{hinton2015KD,SeqKD}. In both cases, the student is supervised on teacher- or data-induced prefixes, whereas at inference it conditions on its own generations, creating a train--inference mismatch.

On-policy methods mitigate this mismatch by querying the teacher on student-generated states, minimizing reverse KL~\citep{MiniLLM}, generalized divergences~\citep{gkd}, or skewed divergences with adaptive replay~\citep{DistiLLM}. Training on student states, however, introduces a new difficulty: when student trajectories are of low quality, teacher supervision on them becomes less reliable. \textsc{SWITCH} therefore lets the teacher intervene in generation when teacher--student discrepancies grow~\citep{SWITCH}, and \textsc{SKD} replaces student tokens that fall outside the teacher's top-$K$~\citep{xu2025speculativeKD}; in \textsc{SKD}, the teacher's ranking decides which tokens enter the trajectory, but the training objective remains distribution matching. Complementary work improves efficiency through difficulty-aware sample selection~\citep{DAKD} and curricula~\citep{l2mkd}. Across these methods, what changes is the states, the divergence, the trajectory construction, or the data schedule; the per-state target remains the teacher's probability distribution. O$^3$PD keeps on-policy states but replaces this target with the teacher's top-$K$ ranking.

\subsection{Distillation Beyond Probability Matching}
\label{sec:related_beyond_probability}

Matching the teacher's distribution requires the student to reproduce
not only which candidates the teacher prefers but also how confident
the teacher is. Relaxations of this target can be characterized by
the transformations of teacher scores that leave the target
unchanged. KL is invariant only to logit shifts. Logit
Standardization~\citep{LogitStandardization} and
\textsc{DIST}~\citep{DIST} additionally remove scale and are invariant
to affine transformations, but still consume continuous scores: a
nonlinear yet order-preserving re-scoring of the teacher changes
their targets. A purely ordinal target is invariant to all strictly
monotone transformations and transfers the teacher's preferences
without its confidence.

Ranking-based methods move toward such ordinal targets but differ from O$^3$PD in either the signal or the setting. On the signal side, RKKD adds a Kendall's $\tau$ loss as an auxiliary to KL, so teacher probabilities still enter the objective~\citep{RKKD}, and PLD, a Plackett--Luce loss akin to ListMLE~\citep{xia2008listmle}, weights each ranked choice by teacher confidence~\citep{Plackett-Luce-Distillation}. On the setting side, these methods, as well as OPKD's pairwise ordering objectives~\citep{OPKD}, are designed for classification over a fixed label set with ground-truth labels, and RankDistil and DGR for ranking fixed pools of items or passages~\citep{RankDistil,DGR}; at student-generated decoding states, by contrast, the candidate set changes from state to state and no ground-truth token exists. Conversely, BiLD operates in LLM distillation and restricts supervision to the teacher's top-$K$ logits, but matches their pairwise differences, which are shift- but not scale-invariant and thus still encode preference strength~\citep{BiLD}. O$^3$PD satisfies both requirements: at each student-generated state, it keeps only the identities and order of the teacher's top-$K$ tokens.

\begin{figure*}[t]
    \centering
    \includegraphics[width=\textwidth]{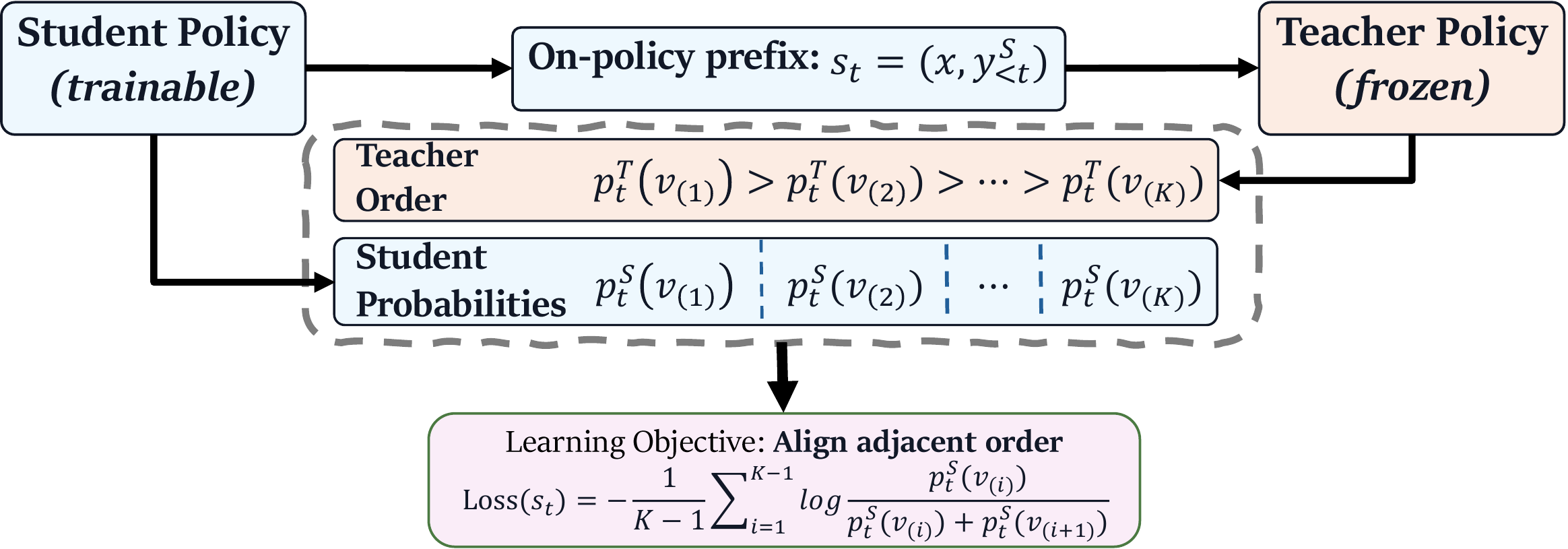}
    \caption{Overview of O$^{3}$PD. The
    figure shows the default adjacent logistic loss,  any other
    order-only objective can take its place (Section~\ref{sec:order_only_principle}). }
    \label{fig:2}
\end{figure*}

\section{Order-Only On-Policy Distillation}
\label{sec:method}

Knowledge distillation can transfer different forms of teacher
supervision, from generated sequences to token-level probability
distributions.
We study an intermediate choice: transferring the teacher's
candidate tokens and their relative ordering without fitting
the teacher's probability magnitudes.
We instantiate this choice as Order-Only On-Policy Distillation
(O$^{3}$PD), which learns from teacher rankings on
student-generated contexts.

\subsection{From Sequence and Distribution Targets to Ordinal Supervision}
\label{sec:distillation_background}

\begin{wraptable}{r}{0.46\textwidth}
\vspace{-20pt}
\centering
\caption{Comparison of KD methods.}
\label{tab:distillation_comparison}
\resizebox{\linewidth}{!}{%
\setlength{\tabcolsep}{4pt}%
\begin{tabular}{@{}lccc@{}}
\toprule
Method & On-policy & \shortstack{Temperature\\invariant} & \shortstack{Magnitude-\\free} \\
\midrule
Sequence KD & \xmark & \xmark & \cmark \\
OPD & \cmark & \xmark & \xmark \\
\textbf{O$^{3}$PD} & \cmark & \cmark & \cmark \\
\bottomrule
\end{tabular}}
\vspace{-8pt}
\end{wraptable}

Sequence-level knowledge distillation (Sequence KD) transfers knowledge through teacher-generated responses. Given a prompt $x$ and a teacher response $y^T$, the student minimizes
\begin{equation}
    \mathcal{L}_{\mathrm{seq}}(x,y^T)
    =
    -\frac{1}{|y^T|}
    \sum_{t=1}^{|y^T|}
    \log \pi_\theta(y_t^T \mid x,y_{<t}^T).
\end{equation}
This objective trains the student to reproduce the teacher's generations, but represents supervision as a selected token at each prefix without explicitly retaining preferences among alternative candidates. Moreover, training follows teacher-generated prefixes rather than output from the current student.

On-policy distillation (OPD) instead obtains teacher supervision on student-generated contexts. The student samples responses from its current policy, and the teacher provides predictive distributions at the resulting prefixes. A representative distribution-matching objective is

\begin{equation}
    \mathcal{L}_{\mathrm{OPD}}(s) = \mathrm{JSD}\big(p^{T}_{K}(\cdot \mid s) \,\|\, \pi_{\theta,K}(\cdot \mid s)\big),
    \label{eq:opd}
\end{equation}

This supervision preserves the teacher's judgments over multiple candidates, but couples their relative ordering with the strength of their assigned probabilities. Even when candidate rankings remain unchanged, different probability assignments yield different training signals. Changing the teacher temperature likewise changes the target distribution while preserving its ordering, introducing an additional temperature-selection problem. When teacher and student differ substantially in reasoning behavior or capability, the teacher's probability assignments may also be difficult for the student to reproduce, complicating knowledge transfer. This motivates a question: can we transfer the teacher's preferences without requiring matching their probability magnitudes?

O$^{3}$PD separates these two forms of information. At each student-generated prefix, we extract the teacher's top-$K$ candidates and their ordering, then train the student through adjacent comparisons. The candidate set determines which tokens are compared, and the ordering determines the preferred direction; teacher probability magnitudes are not fitting targets. This design has three properties. First, it retains \emph{on-policy training}, providing supervision on contexts encountered by the current student. Second, it yields \emph{teacher-temperature invariant supervision}: at a fixed context, positive temperature scaling preserves the candidate set and its ordering, leaving the O$^{3}$PD objective unchanged. This removes the need to select a teacher temperature. More generally, because the objective depends on the teacher only through its top-$K$ ordering, it is invariant to any strictly increasing transformation of the teacher's scores at each context, of which temperature scaling is a special case. Finally, it provides \emph{magnitude-free supervision}: the objective specifies which candidate should be preferred but prescribes no target for how strongly, so the student is never required to reproduce the teacher's probability gaps. Table~\ref{tab:distillation_comparison} summarizes the comparison.

\subsection{Learning through Adjacent Comparisons}
\label{sec:adjacent_alignment}

For a sampled response $y^S$, let $s_t=(x,y^S_{<t})$ denote the context
at position $t$, $z_t^S$ the student logits with
$p_t^S=\operatorname{softmax}(z_t^S)$, and $p_t^T=p^T(\cdot\mid s_t)$ the
teacher distribution. We write
$(v_{t,(1)},\ldots,v_{t,(K)})=\operatorname{TopKOrder}(p_t^T)$ for the
teacher's top-$K$ candidates, in decreasing order of teacher preference.

O$^{3}$PD supervises the student through preferences between adjacent
teacher candidates. For the candidates ranked $i$ and $i+1$, the
student's preference is the probability of selecting the higher-ranked
one when the choice is restricted to the pair:
\begin{equation}
    r_{t,i}^S
    =
    \frac{p_t^S(v_{t,(i)})}
         {p_t^S(v_{t,(i)}) + p_t^S(v_{t,(i+1)})}
    =
    \sigma\!\left(\Delta_{t,i}^S\right),
    \qquad
    \Delta_{t,i}^S
    =
    z_t^S(v_{t,(i)}) - z_t^S(v_{t,(i+1)}).
\end{equation}
Because the softmax normalizer cancels, this Bradley--Terry preference
depends only on the logit difference, and we minimize
$\ell_{t,i}^{\mathrm{adj}}=-\log\sigma(\Delta_{t,i}^S)$. Adjacent
comparisons lose no ordinal information: if all $K-1$ adjacent
inequalities hold, transitivity implies the full teacher ranking, so they
form a minimal description of the order, compared with the $K(K-1)/2$
relations of all-pairs objectives. We average over positions and adjacent
pairs within each response, and then over the minibatch:
\begin{equation}
    \ell(x)
    =
    \frac{1}{|y^S|(K-1)}
    \sum_{t=1}^{|y^S|}
    \sum_{i=1}^{K-1}
    \ell_{t,i}^{\mathrm{adj}},
    \qquad
    \mathcal{L}_{\mathrm{adj}}
    =
    \frac{1}{|\mathcal{B}|}
    \sum_{x\in\mathcal{B}}
    \ell(x).
    \label{eq:batch_loss}
\end{equation}
Sampled responses and teacher rankings are held fixed during
optimization (Algorithm~\ref{alg:o3pd}).

The gradient
$\partial\ell_{t,i}^{\mathrm{adj}}/\partial\Delta_{t,i}^S=-\sigma(-\Delta_{t,i}^S)$
is largest for reversed pairs and decays toward zero once the preferred
order is established. The loss still favors larger margins, but
prescribes no target for them. Distribution matching, by contrast, has
logit gradient $p^S-p^T$, which vanishes only when the student reproduces
the teacher's probabilities exactly.

\subsection{Order-Only as a Distillation Principle}
\label{sec:order_only_principle}

\begin{table}[ht]
\vspace{-12pt}
\centering
\small
\setlength{\tabcolsep}{5pt}
\renewcommand{\arraystretch}{1.3}
\caption{Order-only objectives. $\mathcal{A}$: adjacent pairs;
$\mathcal{P}$: all ordered pairs; $w_{ij}=1/i$.}
\label{tab:order_only_objectives}
\begin{tabular}{@{}lllc@{}}
\toprule
Objective & Comparisons & Structure & Loss \\
\midrule
Adjacent Logistic (O$^{3}$PD) & $\mathcal{A}$, $K-1$ pairs & Pairwise
  & $\mathrm{Avg}_{\mathcal{A}}\!\left[-\log\sigma(\Delta_{ij})\right]$ \\
Logistic & $\mathcal{P}$, $K(K-1)/2$ pairs & Pairwise
  & $\mathrm{Avg}_{\mathcal{P}}\!\left[-\log\sigma(\Delta_{ij})\right]$ \\
Top-Weighted Logistic & $\mathcal{P}$, weighted & Pairwise
  & $\mathrm{Avg}_{\mathcal{P},w}\!\left[-\log\sigma(\Delta_{ij})\right]$ \\
ListMLE & Full list & Listwise
  & $-\frac{1}{K-1}\sum_{i=1}^{K-1}\log\frac{e^{z_i}}{\sum_{j=i}^{K}e^{z_j}}$ \\
\bottomrule
\end{tabular}
\end{table}

The adjacent Bradley--Terry loss is only one way to impose the teacher's
order. Any objective that accesses the teacher only through its top-$K$
candidates and their ordering, and prescribes no target for the student's
score gaps, uses the same supervision as O$^{3}$PD and inherits its
three properties from Section~\ref{sec:distillation_background}: on-policy
training, teacher-temperature invariance, and magnitude-free supervision. Order-only is
thus a principle about what the teacher provides, and O$^{3}$PD is one
instance of it. We examine three further instances that vary which
comparisons are supervised and whether the order is imposed pair by pair
or as a whole (Table~\ref{tab:order_only_objectives}). Dropping the
position index, we write $z_i=z^S(v_{(i)})$ and $\Delta_{ij}=z_i-z_j$, so
that $\Delta_{t,i}^S=\Delta_{i,i+1}$; $\mathcal{P}=\{(i,j):i<j\}$ and
$\mathcal{A}=\{(i,i+1)\}\subset\mathcal{P}$, and
$\operatorname{Avg}_{\mathcal{S},w}$ denotes a $w$-weighted average over
$\mathcal{S}$ (unweighted when $w$ is omitted).

Logistic and Top-Weighted Logistic keep the Bradley--Terry loss~\citep{bradley1952rank} and
change only the comparison set. Logistic supervises all $K(K-1)/2$ pairs,
directly penalizing coarse violations such as the first candidate against
the last, which the adjacent loss constrains only through transitivity.
Top-Weighted Logistic weights each pair by the reciprocal rank of its
preferred candidate, concentrating supervision on the head of the list,
which matters most for decoding. ListMLE instead imposes the
order as a whole: under the Plackett--Luce model~\citep{plackett1975analysis, luce1959individual}, it treats the ranking as
a sequence of choices, each selecting the next candidate from those not
yet ranked. Since Plackett--Luce reduces to Bradley--Terry on any two
candidates, ListMLE expresses the same preferences as O$^{3}$PD, but over
the full list at once. The choice among these objectives therefore
determines how the order is imposed, not what is transferred.

\section{Experiments}

\subsection{Experimental Setup}
\label{sec:experimental_setup}

\paragraph{Models and data.}
We distill within the Qwen3 family~\citep{yang2025qwen3technicalreport}.
Each student (1.7B, 4B, or 8B) is paired with every larger teacher
(4B, 8B, or 14B), giving six teacher--student configurations. The
teacher is kept frozen, and all methods are trained on the
mathematical reasoning subset of OpenThoughts~\citep{openthoughts}.
Full training details are given in Appendix~\ref{app:setup}.

\paragraph{Evaluation.}
We evaluate on three competition-level benchmarks, HMMT25, AIME24,
and AIME25, and report Avg@12: the accuracy of the final answer
averaged over 12 sampled responses per problem, drawn as four samples
under each of three seeds. Average denotes the unweighted mean over
the three benchmarks, and all differences are reported in absolute
points. For OPD and O$^{3}$PD, we evaluate every 20 training steps and
report the checkpoint with the highest Average. This selection can
only favor a method whose performance declines during training. For
the 8B$\rightarrow$1.7B pair, OPD peaks early and then degrades while
O$^{3}$PD keeps improving (Figure~\ref{fig:curves}), so comparing
final checkpoints instead would widen the gap in favor of O$^{3}$PD.

\paragraph{Baselines.}
We compare O$^{3}$PD with the base student, SFT, GRPO~\citep{GRPO},
and OPD~\citep{MiniLLM,gkd}. SFT and GRPO results are taken from
\citet{OPSD}, which uses the same student models, training data, and
evaluation protocol. OPD is our primary baseline and is trained under
exactly the same configurations as O$^{3}$PD. It minimizes the
Jensen--Shannon divergence between the student and teacher
distributions restricted to the teacher's top-$K$ tokens, with $K=8$
and teacher temperature $T=1$, so that both methods see the same
candidate set at every prefix and differ only in whether the probability values enter the loss.

\subsection{Main Results}
\label{sec:main_results}

\begin{table*}[t]
\centering
\small
\caption{Main results on mathematical reasoning (Avg@12, \%).
Values in parentheses are differences relative to OPD with the same
teacher--student pair. Gray rows are non-distillation
references; SFT and GRPO results are taken from \citet{OPSD}.}
\label{tab:o3pd-main}

\definecolor{gainGreen}{RGB}{100,170,125}
\definecolor{dropOrange}{RGB}{225,160,95}
\definecolor{baselinegray}{gray}{0.93}

\def\gain#1{\textcolor{gainGreen}{(+#1)}}
\def\drop#1{\textcolor{dropOrange}{($-#1$)}}
\def\graycell#1{\cellcolor{baselinegray}#1}

\setlength{\tabcolsep}{5pt}
\renewcommand{\arraystretch}{1.0}

\begin{tabular}{@{}cclllll@{}}
\toprule
Student & Teacher & Method & HMMT25 & AIME24 & AIME25 & Average \\
\midrule

\multirow{5}{*}{Qwen3-8B}
& -- & \graycell{Base}
& \graycell{44.4} & \graycell{75.8}
& \graycell{66.4} & \graycell{62.2} \\
& -- & \graycell{SFT}
& \graycell{42.9} & \graycell{72.3}
& \graycell{64.2} & \graycell{59.8} \\
& -- & \graycell{GRPO}
& \graycell{46.7} & \graycell{76.4}
& \graycell{68.9} & \graycell{64.0} \\
& \multirow{2}{*}{Qwen3-14B} & OPD
& 46.1 & 75.8 & 70.0 & 64.0 \\
& & O$^3$PD
& 46.7 \gain{0.6}
& 76.7 \gain{0.9}
& 70.3 \gain{0.3}
& 64.6 \gain{0.6} \\

\midrule

\multirow{7}{*}{Qwen3-4B}
& -- & \graycell{Base}
& \graycell{43.9} & \graycell{75.8}
& \graycell{64.7} & \graycell{61.5} \\
& -- & \graycell{SFT}
& \graycell{43.4} & \graycell{70.2}
& \graycell{62.3} & \graycell{58.6} \\
& -- & \graycell{GRPO}
& \graycell{44.4} & \graycell{75.6}
& \graycell{68.1} & \graycell{62.7} \\
& \multirow{2}{*}{Qwen3-14B} & OPD
& 42.2 & 74.4 & 66.1 & 60.9 \\
& & O$^3$PD
& 45.0 \gain{2.8}
& 74.2 \drop{0.2}
& 70.3 \gain{4.2}
& 63.2 \gain{2.3} \\
\cmidrule{2-7}
& \multirow{2}{*}{Qwen3-8B} & OPD
& 43.1 & 73.9 & 66.9 & 61.3 \\
& & O$^3$PD
& 44.2 \gain{1.1}
& 74.7 \gain{0.8}
& 70.6 \gain{3.7}
& 63.2 \gain{1.9} \\

\midrule

\multirow{9}{*}{Qwen3-1.7B}
& -- & \graycell{Base}
& \graycell{24.4} & \graycell{53.1}
& \graycell{35.6} & \graycell{37.7} \\
& -- & \graycell{SFT}
& \graycell{22.7} & \graycell{48.4}
& \graycell{36.3} & \graycell{35.8} \\
& -- & \graycell{GRPO}
& \graycell{23.7} & \graycell{51.1}
& \graycell{38.3} & \graycell{37.7} \\
& \multirow{2}{*}{Qwen3-14B} & OPD
& 25.6 & 52.2 & 39.4 & 39.1 \\
& & O$^3$PD
& 26.1 \gain{0.5}
& 51.4 \drop{0.8}
& 40.8 \gain{1.4}
& 39.4 \gain{0.3} \\
\cmidrule{2-7}
& \multirow{2}{*}{Qwen3-8B} & OPD
& 27.2 & 49.4 & 37.8 & 38.1 \\
& & O$^3$PD
& 27.2 (0.0)
& 51.9 \gain{2.5}
& 43.1 \gain{5.3}
& 40.7 \gain{2.6} \\
\cmidrule{2-7}
& \multirow{2}{*}{Qwen3-4B} & OPD
& 27.8 & 48.9 & 37.2 & 38.0 \\
& & O$^3$PD
& 24.7 \drop{3.1}
& 55.3 \gain{6.4}
& 42.8 \gain{5.6}
& 40.9 \gain{2.9} \\

\bottomrule
\end{tabular}
\end{table*}

\paragraph{O$^{3}$PD consistently outperforms baselines.}
Table~\ref{tab:o3pd-main} shows that O$^{3}$PD achieves a higher
average than OPD in all six configurations. The gain is modest for the
8B student (+0.6) and larger for the smaller students: +2.3 and +1.9
for the 4B student (14B and 8B teachers), and +2.6 and +2.9 for the
1.7B student with the 8B and 4B teachers. The exception is the
14B$\rightarrow$1.7B pair, which improves by only 0.3, suggesting that
when the capacity gap is largest, even the teacher's ranking becomes
harder for the student to reproduce. The gains concentrate on AIME.
O$^{3}$PD improves AIME25 in every configuration, by 3.4 points on
average and up to 5.6. It also gives the largest single gain in the
table, +6.4 on AIME24 for 4B$\rightarrow$1.7B. Overall, O$^{3}$PD
improves over OPD in 14 of the 18 comparisons.
O$^{3}$PD also exceeds the base model, SFT, and GRPO in average score
in every configuration. For the 1.7B student, for example, it reaches
40.9, compared with 37.7 for GRPO and 35.8 for SFT. SFT falls below the
base model at every scale, which highlights the value of on-policy
supervision.

\begin{figure*}[t]
    \centering
    \includegraphics[width=\textwidth]{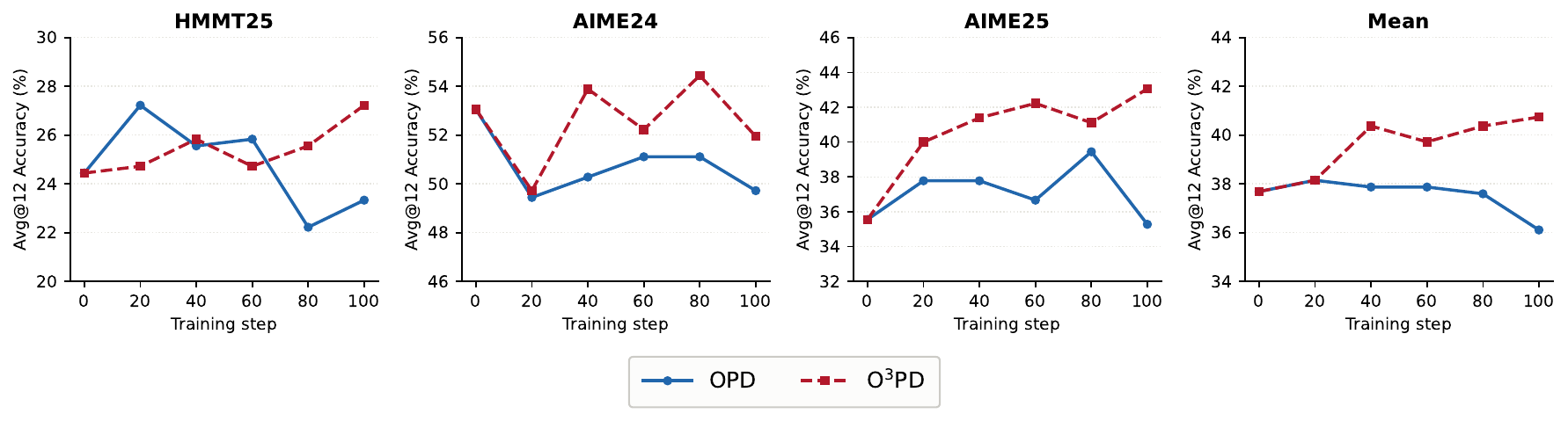}
    \caption{Performance curve of OPD and O$^{3}$PD for the
    8B$\rightarrow$1.7B pair. Avg@12 is evaluated every 20 training
    steps; step 0 is the base student.}
    \label{fig:curves}
\end{figure*}

\paragraph{O$^{3}$PD improves steadily throughout training.}
Figure~\ref{fig:curves} tracks both methods for the 8B$\rightarrow$1.7B
pair. OPD peaks at step 20 (38.1) and then declines below the base
student (37.7). O$^{3}$PD instead keeps improving and reaches its best
mean (40.7) at the final step. The contrast is sharpest on AIME25, where
O$^{3}$PD climbs from 35.6 to 43.1 while OPD falls back to its starting
point. The gains of O$^{3}$PD therefore do not depend on checkpoint
selection: at the end of training, the gap between the two methods
exceeds the 2.6-point gap between their best checkpoints.

\paragraph{Smaller teachers suffice.}
O$^{3}$PD with a smaller teacher outperforms OPD with the 14B teacher,
both for the 1.7B student (40.9 with the 4B teacher vs.\ 39.1) and for
the 4B student (63.2 with the 8B teacher vs.\ 60.9). O$^{3}$PD thus
achieves stronger distillation with a cheaper teacher.

\textbf{O$^3$PD removes temperature tuning.}
For OPD, the teacher temperature is a consequential hyperparameter. Appendix Figure~\ref{fig:heat} sweeps $T \in \{0.25, 0.5, 1, 2, 4, 8\}$ for every teacher--student pair, together with an additional Qwen3-0.6B student. No single value is best across pairs: the optimum ranges from $T=0.25$ for 8B$\rightarrow$1.7B to $T=2$ for 14B$\rightarrow$8B and 4B$\rightarrow$1.7B. Even after only three training steps, the choice of $T$ alone moves HMMT25 accuracy by up to 6.4 points for students of 1.7B and above (8B$\rightarrow$4B: 44.7 at $T=1$ vs.\ 38.3 at $T=8$). Sensitivity grows as the student shrinks, and the 0.6B student collapses at high temperatures: 14B$\rightarrow$0.6B falls from 11.9 at $T=1$ to 1.1 at $T=8$, and 1.7B$\rightarrow$0.6B drops to 0.3. Because dividing the teacher logits by any $T>0$ leaves the top-$K$ ordering unchanged, O$^3$PD receives identical targets at every temperature.

\subsection{Ablation Study}
\label{sec:ablation}

We ablate the ranking objective and the candidate-set size $K$ on the
8B$\rightarrow$1.7B pair, training each variant for 100 steps
(Table~\ref{tab:o3pd-objective-k}); further details are in
Appendix~\ref{app:setup}.

\paragraph{Ranking objective.}
Besides Adjacent Logistic, we evaluate the Logistic, Top-Weighted
Logistic, and ListMLE variants of O$^{3}$PD. All variants outperform
OPD at $K=8$, by 1.3 to 2.6 points, so ordinal supervision is robust
to the choice of ranking loss. Adjacent Logistic and ListMLE perform
best, within 0.1 points of each other, even though Adjacent Logistic
supervises only 7 of the 28 pairs used by Logistic. By transitivity,
the adjacent relations already fix the full ranking, so supervising
the remaining pairs brings no further gain. We use Adjacent Logistic
by default, as it matches the best result with the fewest constraints.

\newsavebox{\ablationbox}

\begin{figure}[t]
\centering
\sbox{\ablationbox}{%
  \resizebox{0.56\textwidth}{!}{%
  \setlength{\tabcolsep}{4pt}%
  \begin{tabular}{@{}lcccc@{}}
  \toprule
  Method & HMMT25 & AIME24 & AIME25 & Avg. \\
  \midrule
  OPD ($K{=}8$) & 27.2 & 49.4 & 37.8 & 38.1 \\
  \midrule
  \multicolumn{5}{@{}l}{\textit{Adjacent, varying $K$}} \\
  \quad $K{=}4$  & 26.1 & 53.1 & 40.8 & 40.0 \\
  \quad $K{=}8$  & 27.2 & 51.9 & \textbf{43.1} & \textbf{40.7} \\
  \quad $K{=}16$ & 24.7 & 51.9 & 42.5 & 39.7 \\
  \midrule
  \multicolumn{5}{@{}l}{\textit{Other objectives, $K{=}8$}} \\
  \quad Logistic     & 25.6 & 53.3 & 41.7 & 40.2 \\
  \quad Top-Weighted & 24.2 & 53.3 & 40.6 & 39.4 \\
  \quad ListMLE      & 26.4 & \textbf{53.6} & 41.7 & 40.6 \\
  \bottomrule
  \end{tabular}}}%
\begin{minipage}[b]{\wd\ablationbox}
  \centering
  \captionof{table}{Ablations on the 8B$\rightarrow$1.7B pair (100 steps).
  Adjacent with $K{=}8$ is the default O$^{3}$PD setting; Top-Weighted
  denotes Top-Weighted Logistic.}
  \label{tab:o3pd-objective-k}
  \vspace{4pt}
  \raisebox{\dp\ablationbox}{\usebox{\ablationbox}}
\end{minipage}%
\hfill
\raisebox{33pt}{%
\begin{minipage}[t]{\dimexpr\textwidth-\wd\ablationbox-1.2em\relax}
  \centering
  \includegraphics[width=\linewidth,
    height=\dimexpr\ht\ablationbox+\dp\ablationbox\relax,
    keepaspectratio]{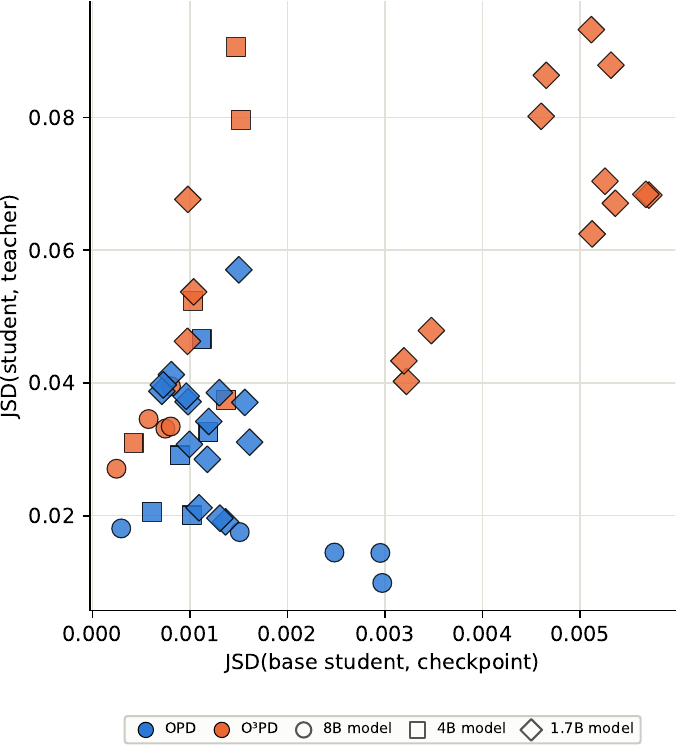}
  \captionof{figure}{Training checkpoints: JSD to the base student
  vs.\ JSD to the teacher.}
  \label{fig:drift}
\end{minipage}}
\end{figure}

\paragraph{Candidate-set size.}
All three sizes outperform OPD, by 1.6 to 2.6 points, and $K=8$
performs best. Neither a shorter ($K=4$) nor a longer ($K=16$) list
falls far behind, so O$^{3}$PD is not sensitive to the exact choice
of $K$ within this range.

\section{Analysis}
\label{sec:analysis}

Section~\ref{sec:main_results} shows that O$^{3}$PD produces stronger
students than OPD and, on the 8B$\rightarrow$1.7B pair, keeps
improving where OPD degrades. We now examine where this advantage
comes from. We first ask what the student acquires from the teacher
under ordinal supervision (Section~\ref{sec:analysis_order}), and then
how training reshapes the student relative to its initialization
(Section~\ref{sec:analysis_drift}).

\subsection{Learning the Teacher's Order, Not Its Probabilities}
\label{sec:analysis_order}

\begin{figure*}[t]
    \centering
    \includegraphics[width=\textwidth]{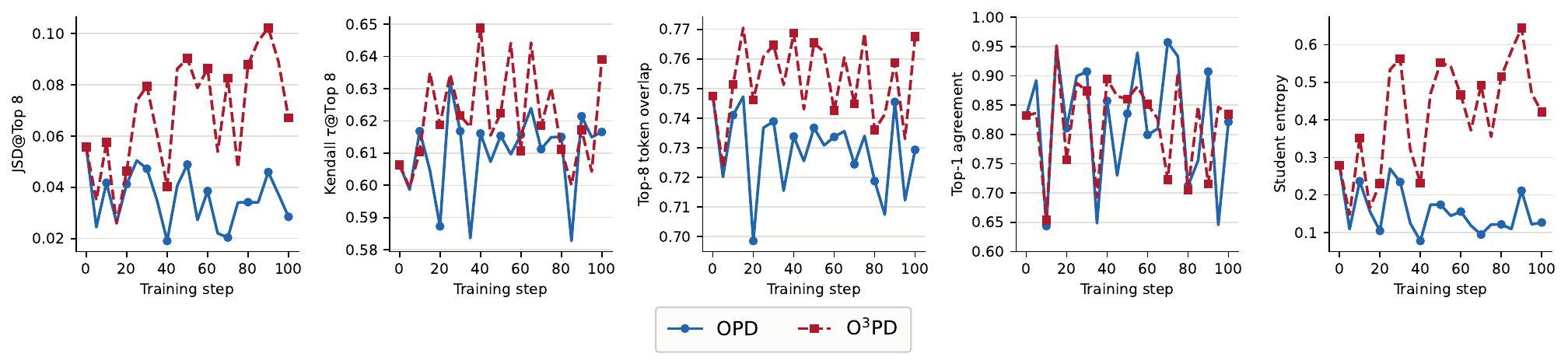}
    \caption{Token-level agreement between student and teacher during
    training for the 8B$\rightarrow$1.7B pair, computed on
    student-generated responses. From left to right: JSD@Top-8,
    Kendall's $\tau$, top-8 overlap, top-1 agreement, and student
    entropy. O$^{3}$PD departs further from the teacher's
    probabilities than OPD, yet agrees more closely with its ordering
    and candidate set, and maintains higher entropy throughout
    training.}
    \label{fig:alignment}
\end{figure*}

We measure token-level agreement between the student and the teacher
on student-generated responses for the 8B$\rightarrow$1.7B pair
(Figure~\ref{fig:alignment}). At each position $t$, let
$\mathcal{T}^T_t$ and $\mathcal{T}^S_t$ denote the top-8 tokens of the
teacher and the student. \emph{JSD@Top-8} is the Jensen--Shannon
divergence between the two distributions restricted to
$\mathcal{T}^T_t$ and renormalized, and measures how closely the
student reproduces the teacher's probability values. \emph{Kendall's
$\tau$} compares the orderings that the two models induce on
$\mathcal{T}^T_t$, and \emph{top-8 overlap}
$|\mathcal{T}^T_t \cap \mathcal{T}^S_t|/8$ measures whether the
student considers the same candidates plausible. \emph{Top-1
agreement} is the fraction of positions at which the two models share
the most likely token, and \emph{student entropy} is the entropy of
the student's next-token distribution over the full vocabulary. All
metrics are averaged over positions.

The first metric concerns probability values. OPD behaves as its
objective prescribes and keeps the JSD to the teacher low throughout
training. The JSD of O$^{3}$PD is consistently higher and grows as
training proceeds, which is expected since O$^{3}$PD never asks the
student to match the teacher's probability values. On ordering and
candidate set, the relation reverses. O$^{3}$PD achieves a higher
Kendall's $\tau$ than OPD for most of training and ends clearly above
it, and its top-8 overlap stays above that of OPD at nearly every
evaluation, while top-1 agreement is comparable. O$^{3}$PD thus
transfers the structure that governs generation, namely which
candidates the teacher considers plausible and how it orders them,
and leaves the probability values free. The student also keeps a
broader distribution: its entropy stays above that of OPD throughout
training, whereas probability matching pulls the student toward the
teacher's confidence profile. Together with Figure~\ref{fig:curves},
these results show that closer probability matching does not
translate into better reasoning: OPD attains the lower JSD, yet
O$^{3}$PD attains the higher accuracy.

\subsection{Reshaping the Student}
\label{sec:analysis_drift}

Figure~\ref{fig:drift} places each checkpoint according to its
distance from the base student and its distance from the teacher.
For every student size, O$^{3}$PD checkpoints sit further from the
teacher's probabilities than OPD checkpoints, consistent with
Section~\ref{sec:analysis_order}. How far the student moves from its
initialization, however, depends on student size. For the 1.7B
student, OPD checkpoints stay close to the base student, whereas
O$^{3}$PD checkpoints move several times further away. For the 4B
student, the two methods move comparable distances, and for the 8B
student, OPD moves further than O$^{3}$PD. Ordinal supervision thus
does not uniformly produce larger updates; it produces them for the
smallest student, where matching the teacher's probabilities is
hardest.

The mechanism is intuitive. A small student cannot reproduce the
probability values of a much larger teacher, so a magnitude-matching
objective spends much of its capacity on a target that the student
cannot reach. Ordinal supervision removes this constraint. The student
is free to restructure its own distribution, provided that it
preserves the teacher's ordering of the candidates, and the smallest
student makes the most use of this freedom.
\section{Conclusion}
\label{sec:conclusion}

We presented Order-Only On-Policy Distillation (O$^{3}$PD), which
trains the student to reproduce the teacher's top-$K$ token ranking
without matching the teacher's probability values. Because the
supervision depends only on the teacher's order, O$^{3}$PD is
invariant to the teacher temperature and requires no temperature
tuning. Across six Qwen3 teacher--student configurations, O$^{3}$PD
outperforms OPD on the three-benchmark average in every configuration,
also exceeds SFT and GRPO, and, on the 8B$\rightarrow$1.7B pair, keeps
improving where OPD degrades. Our analysis shows that O$^{3}$PD
reproduces the teacher's ranking and candidate set more faithfully
than OPD while departing further from its probability values, and
that it allows the smallest student to move substantially further
from its initialization. The order-only principle is not tied to a
single loss: other ranking objectives such as ListMLE also perform well. These results answer the question posed at the outset: for
on-policy distillation, the teacher's ordering is sufficient, and
matching its probability values is not only unnecessary but can hold
the student back.

\section*{Ethics statement}

This work studies training objectives for knowledge distillation and does not involve human subjects, user data, or personally identifiable information. All experiments use publicly released models from the Qwen3 family and publicly available data: the mathematical reasoning subset of OpenThoughts for training, and the HMMT25, AIME24, and AIME25 competition problems for evaluation. We use these resources in accordance with their respective licenses and do not release any new datasets.  Our experiments are restricted to mathematical reasoning, so we do not evaluate how ordinal distillation affects safety alignment or other behaviors in open-ended generation. We are not aware of any conflicts of interest related to this work.
\section*{Reproducibility statement}

We have made several efforts to ensure the reproducibility of our results. The O$^{3}$PD objective is fully specified in Section~\ref{sec:method} (Equations~(3) and~(4)), and one training iteration is given as pseudocode in Algorithm~\ref{alg:o3pd}. Appendix~\ref{app:setup} describes the experimental setup: Table~\ref{tab:hyperparams} lists all training and evaluation hyperparameters, including the LoRA configuration, rollout length, candidate-set size, decoding parameters, and random seeds, and Figure~\ref{fig:prompt} gives the exact prompt template used for both training rollouts and evaluation. The OPD baseline shares the same setup. All models (Qwen3), training data (OpenThoughts), and evaluation benchmarks (HMMT25, AIME24, AIME25) are publicly available, and our evaluation protocol (Avg@12 over three seeds, answer extraction from \texttt{\textbackslash boxed\{\}}) is described in Section~\ref{app:setup}. 

\section*{Acknowledgments}
This work was supported by the National Science Foundation (NSF) under award CNS-2318210, which also funded the iTiger GPU cluster~\cite{itiger} used for our experiments.

\bibliography{iclr2027_conference}
\bibliographystyle{iclr2027_conference}

\appendix
\section{Experimental Setup}
\label{app:setup}

\subsection{Hyperparameters}

Table~\ref{tab:hyperparams} lists the hyperparameters used for
training and evaluation. We fine-tune the student with LoRA and train
for 100 steps with an effective batch size of 32. During training,
the teacher runs in thinking mode while the student runs in
non-thinking mode, and each student rollout is limited to 1{,}024
tokens. At every student-generated prefix, both OPD and O$^{3}$PD use
the teacher's top-8 candidates. For evaluation, we checkpoint every
20 steps and report Avg@12, obtained by drawing four samples per
problem under each of three seeds.

\begin{table}[h]
\centering
\caption{Hyperparameters for O$^{3}$PD and OPD training and evaluation.
All methods share the same settings unless noted otherwise.}
\label{tab:hyperparams}
\small
\begin{tabular}{@{}ll@{}}
\toprule
\textbf{Hyperparameter} & \textbf{Value} \\
\midrule
\multicolumn{2}{@{}l}{\textit{Training}} \\
Student fine-tuning         & LoRA ($r=64$, $\alpha=128$) \\
Learning Rate & 5e-6 \\
Teacher thinking mode       & Enabled \\
Student thinking mode       & Disabled \\
Rollout length              & 1{,}024 tokens \\
Effective batch size        & 32 \\
Training steps              & 100 \\
Training seed               & 42 \\
Top-$K$ candidates (OPD and O$^{3}$PD) & 8 \\
Ranking objective (O$^{3}$PD)  & Adjacent logistic \\
Teacher temperature (OPD and O$^{3}$PD)   & 1 \\
\midrule
\multicolumn{2}{@{}l}{\textit{Evaluation}} \\
Evaluation interval         & Every 20 steps \\
Evaluation seeds            & 41, 42, 43 \\
Samples per problem         & 4 per seed (12 total, Avg@12) \\
Sampling temperature        & 1.0 \\
Top-$p$                     & 1.0 \\
Top-$k$                     & $-1$ (disabled) \\
Min-$p$                     & 0 \\
Presence penalty            & 0 \\
Max new tokens              & 38{,}912 \\
Max model length            & 40{,}960 \\
\bottomrule
\end{tabular}
\end{table}

\subsection{Prompt Template}

All training rollouts and evaluations use the same prompt, shown in
Figure~\ref{fig:prompt}. The prompt asks the model to reason step by
step and to place its final answer in \texttt{\textbackslash boxed\{\}},
from which we extract the answer for scoring.

\begin{figure}[h]
\centering
\begin{tcolorbox}[
    colback=gray!5,
    colframe=gray!60,
    boxrule=0.6pt,
    arc=2pt,
    left=6pt, right=6pt, top=4pt, bottom=4pt,
    width=0.92\textwidth,
    title=\textbf{Prompt template},
    fonttitle=\small,
    coltitle=black,
    colbacktitle=gray!20
]
\small\ttfamily
Problem: \textit{\{problem\}}\\[4pt]
Reason step by step and put the final answer in \textbackslash boxed\{\}.
\end{tcolorbox}
\caption{Prompt template used for training rollouts and evaluation.
\textit{\{problem\}} is replaced with the problem statement.}
\label{fig:prompt}
\end{figure}

\subsection{Training Procedure}

Algorithm~\ref{alg:o3pd} summarizes one training iteration of
O$^{3}$PD. For each prompt, the student samples a response, and the
teacher is queried at every prefix of that response to obtain its
top-$K$ tokens in descending order of probability. The student is then
updated with the adjacent ranking loss, which averages the pairwise
logistic terms over all positions and all $K-1$ adjacent pairs. The
teacher's probabilities are used only to select and order the
candidates and do not appear in the loss.

\begin{algorithm}[t]
\caption{Order-Only On-Policy Distillation (O$^{3}$PD)}
\label{alg:o3pd}
\begin{algorithmic}[1]
\Require Prompt set $\mathcal{D}$; frozen teacher $p^T$;
student policy $\pi_\theta$; top-$K$ size $K$
\While{not converged}
\State Sample a minibatch $\mathcal{B}\subset\mathcal{D}$
\ForAll{$x\in\mathcal{B}$}
\State Sample an on-policy response
$y^S\sim\pi_\theta(\cdot\mid x)$
\State At each $s_t=(x,y^S_{<t})$, obtain
\[
(v_{t,(1)},\ldots,v_{t,(K)})
\gets
\operatorname{TopKOrder}\!\left(p_t^T(\cdot)\right)
\]
\State Compute the adjacent ranking loss
\[
\ell(x)
\gets
-\frac{1}{|y^S|(K-1)}
\sum_{t=1}^{|y^S|}
\sum_{i=1}^{K-1}
\log
\frac{
p_t^S(v_{t,(i)})
}{
p_t^S(v_{t,(i)})
+
p_t^S(v_{t,(i+1)})
}
\]
\EndFor
\State $\displaystyle
\mathcal{L}_{\mathrm{adj}}
\gets
\frac{1}{|\mathcal{B}|}
\sum_{x\in\mathcal{B}}\ell(x)$;
update $\theta$
\EndWhile
\end{algorithmic}
\end{algorithm}

\subsection{Infrastructure}

All models are trained with Transformers~\cite{wolf2020transformers} and evaluated using vLLM~\cite{kwon2023vllm} on NVIDIA H100 GPUs. 

\section{Teacher Temperature Sensitivity of OPD}
\label{app:temperature}

\begin{figure}[h]
    \centering
    \includegraphics[width=\textwidth]{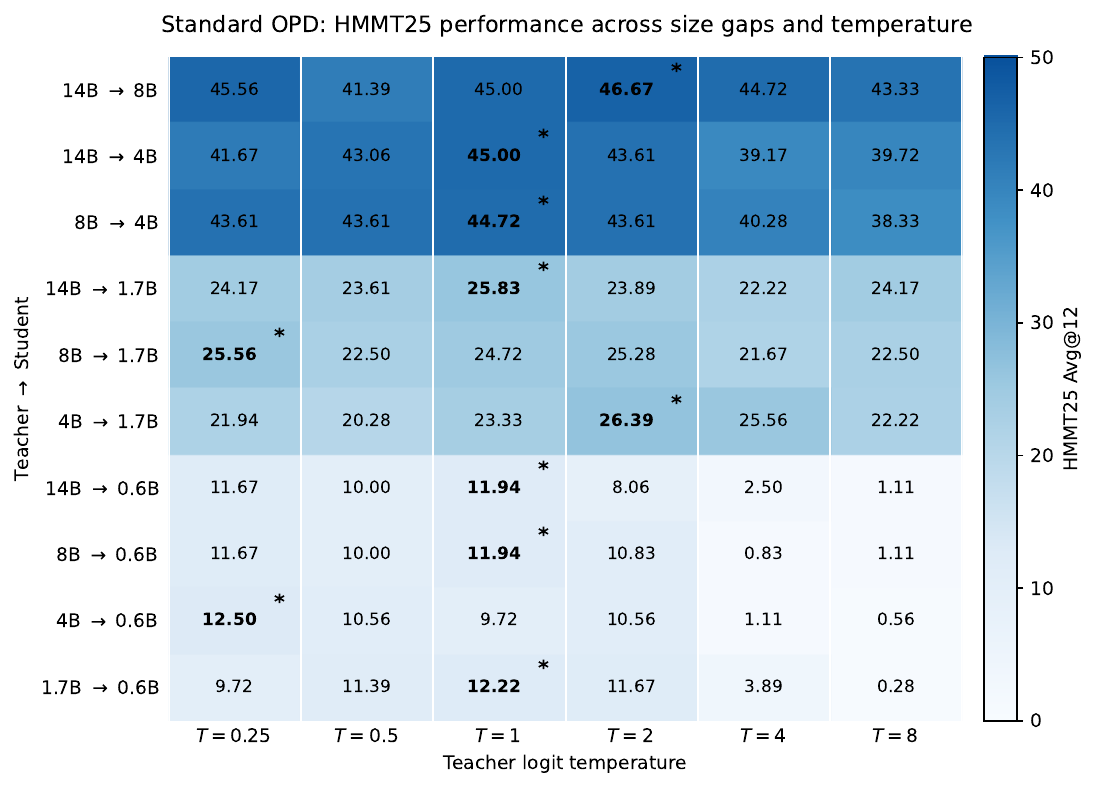}
    \caption{HMMT25 Avg@12 of standard OPD across teacher--student
    pairs (rows) and teacher logit temperatures $T$ (columns). Each run
    is trained for only 3 steps with a batch size of 32. Bold values
    marked with $*$ indicate the best temperature for each pair.}
    \label{fig:heat}
\end{figure}

Figure~\ref{fig:heat} reports the teacher-temperature sweep. We train standard OPD for every
teacher--student pair at each $T\in\{0.25,0.5,1,2,4,8\}$ and evaluate
HMMT25 Avg@12. The sweep covers the six configurations in
Table~\ref{tab:o3pd-main} and additionally includes a Qwen3-0.6B
student.

OPD is highly sensitive to $T$, and no single temperature is best
across pairs: the optimum ranges from $T=0.25$ (8B$\rightarrow$1.7B)
to $T=2$ (14B$\rightarrow$8B, 4B$\rightarrow$1.7B). Sensitivity
increases as the student becomes smaller. For the 0.6B student,
performance collapses at high temperatures, e.g., 14B$\rightarrow$0.6B
falls from 11.94 at $T=1$ to 1.11 at $T=8$. O$^{3}$PD avoids this
selection problem entirely, since its ranking targets are identical
for every $T>0$.

\end{document}